\documentclass[11pt]{article}

\usepackage[final]{acl}

\usepackage{times}
\usepackage{latexsym}
\usepackage{amsmath}
\usepackage{amssymb}
\usepackage{soul}       
\usepackage{xcolor}     
\newsavebox{\wfbox}
\newcommand{\wrongflipbox}[3]{
  \begingroup
  \setlength{\fboxsep}{4pt}%
  \setlength{\fboxrule}{1.2pt}%
  \fcolorbox{#1}{#2}{\parbox{0.88\columnwidth}{#3}}%
  \endgroup
}

\usepackage[T1]{fontenc}

\usepackage[utf8]{inputenc}

\usepackage{microtype}

\usepackage{inconsolata}

\usepackage{graphicx}

\title{FregeLogic at SemEval 2026 Task 11: A Hybrid Neuro-Symbolic Architecture for Content-Robust Syllogistic Validity Prediction}

\author{Adewale Akinfaderin \\
  Amazon Web Services \\
  Seattle, WA\\
  \texttt{akinfaa@amazon.com} \\\And
  Nafi Diallo \\
  Amazon Web Services \\
  Seattle, WA\\
  \texttt{nafid@amazon.com} \\}

\begin{document}
\maketitle

\begin{abstract}
We present FregeLogic, a hybrid neuro-symbolic system for SemEval-2026 Task 11 (Subtask 1), which addresses syllogistic validity prediction while reducing content effects on predictions. Our approach combines an ensemble of five LLM classifiers, spanning three open-weights models (Llama 4 Maverick, Llama 4 Scout, and Qwen3-32B) paired with varied prompting strategies, with a Z3 SMT solver that serves as a formal logic tiebreaker. The central hypothesis is that LLM disagreement within the ensemble signals likely content-biased errors, where real-world believability interferes with logical judgment. By deferring to Z3's structurally-grounded formal verification on these disputed cases, our system achieves 94.3\% accuracy with a content effect of 2.85 and a combined score of 41.88 in nested 5-fold cross-validation on the dataset ($N{=}960$). This represents a 2.76-point improvement in combined score over the pure ensemble (39.12), with a 0.9\% accuracy gain, driven by a 16\% reduction in content effect (3.39 $\to$ 2.85). Adopting structured-output API calls for Z3 extraction reduced failure rates from ${\sim}22\%$ to near zero, and an Aristotelian encoding with existence axioms was validated against task annotations. Our results suggest that targeted neuro-symbolic integration, applying formal methods precisely where ensemble consensus is lowest, can improve the combined accuracy-plus-content-effect metric used by this task.
\end{abstract}

\section{Introduction}

Syllogistic reasoning is a fundamental form of deductive inference studied extensively in logic and cognitive science \citep{eisape2024systematic, bertolazzi2024systematic}. A key challenge in evaluating reasoning capabilities of language models is \textit{content effects}: the tendency for real-world believability to interfere with purely logical judgment \citep{dasgupta2022language}. Mechanistic analyses have shown that language models develop reasoning circuits during pre-training, but these are susceptible to contamination from world knowledge \citep{kim-etal-2025-reasoning}, a phenomenon documented across tasks, model families, and domains \citep{wysocka2025syllobio, ozeki2024exploring}. SemEval-2026 Task 11 \citep{valentino-etal-2026-semeval} formalizes this challenge by evaluating systems on syllogistic validity prediction using a combined metric that rewards both accuracy and low content effect.

In this paper, we describe FregeLogic, a hybrid neuro-symbolic system that exploits the complementary strengths of LLM ensembles and formal logic solvers. We construct a five-member ensemble from three open-weight model families (Llama~4 Maverick, Llama~4 Scout, Qwen3-32B) paired with varied prompting strategies, then use the Z3 SMT solver \citep{demoura2008z3} as a tiebreaker when the ensemble produces a narrow 3--2 vote split. The design is motivated by an empirical observation: close votes in the ensemble disproportionately coincide with content-biased errors, precisely the cases where a content-neutral formal verifier can add value \citep{bayless2025neurosymbolic, ranaldi2025improving}. In nested 5-fold cross-validation on the dataset ($N{=}960$), this selective intervention reduces content effect by 16\% (3.39 $\to$ 2.85) while improving accuracy by 0.9\%, yielding a combined score of 41.88 compared to 39.12 for the pure ensemble.

\section{Background}

\subsection{Task Description and Metrics}

SemEval-2026 Task 11 \citep{valentino-etal-2026-semeval} evaluates syllogistic reasoning while disentangling logical structure from semantic content. We participate in Subtask~1: binary classification of syllogisms as valid or invalid. The dataset comprises 960 syllogisms annotated with validity labels and plausibility metadata (believable/unbelievable), balanced across four subgroups.

The task uses a combined score that jointly rewards high accuracy and low \textbf{content effect} (CE), which measures how much a system's predictions are influenced by believability rather than logical structure. The combined score applies a logarithmic penalty for content bias: $\text{Score} = \text{Accuracy} / (1 + \ln(1 + \text{CE}))$. A content effect of zero indicates predictions uninfluenced by believability. Full metric definitions are in \citet{valentino-etal-2026-semeval}.

\subsection{Related Work}

\paragraph{Content effects in LLMs.} \citet{dasgupta2022language} demonstrated that large language models exhibit human-like content effects across syllogism validity judgments and other reasoning tasks. \citet{eisape2024systematic} showed that even the largest models in the PaLM 2 family exhibit systematic biases including sensitivity to variable ordering, a structural bias related to, but distinct from, semantic content effects. \citet{bertolazzi2024systematic} found that pre-trained LLM behavior can be explained by heuristics from cognitive science, and \citet{ozeki2024exploring} confirmed that LLMs exhibit human-like reasoning biases with primary limitations in the reasoning process itself.

\paragraph{Mechanistic interpretations.} \citet{kim-etal-2025-reasoning} uncovered a three-stage reasoning circuit for syllogistic inference involving middle-term suppression and information propagation via mover heads. Critically, this circuit is susceptible to contamination from belief biases encoded in additional attention heads, providing mechanistic evidence for content effects and motivating our use of formal verification.

\paragraph{Neuro-symbolic approaches.} \citet{ranaldi2025improving} proposed quasi-symbolic chain-of-thought prompting to improve robustness on reasoning tasks. \citet{bayless2025neurosymbolic} presented a neurosymbolic framework for verifying logical correctness using SMT solvers, and \citet{akinfaderin2026verafi} applied similar ideas to financial AI. \citet{wysocka2025syllobio} showed that zero-shot LLMs achieve between 23\% and 70\% on biomedical syllogistic reasoning. \citet{valentino2025mitigating} and \citet{maraia2026abstract} explored activation-level approaches to mitigate content effects. Our work differs from pure-LLM approaches by introducing structurally-grounded formal verification, from pure-formal approaches by handling paraphrase and extraction failures gracefully, from \citet{bayless2025neurosymbolic} by applying verification selectively only where ensemble consensus is low, and from activation-steering approaches \citep{valentino2025mitigating, maraia2026abstract} by requiring no access to model internals.

\section{System Overview}

\begin{figure}[t]
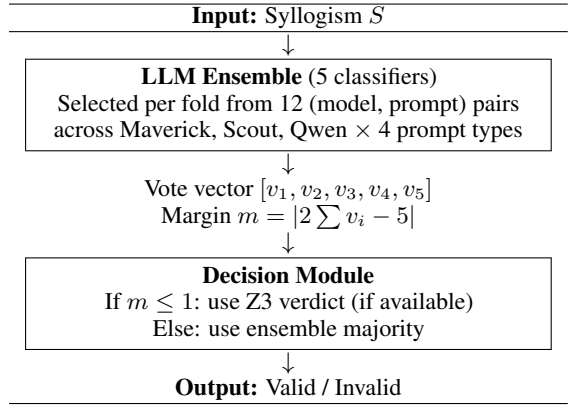

\centering
\small
\begin{tabular}{c}
\hline
\textbf{Input:} Syllogism $S$ \\
\hline
$\downarrow$ \\
\framebox[0.9\columnwidth]{%
\parbox{0.85\columnwidth}{\centering
\textbf{LLM Ensemble} (5 classifiers)\\
Selected per fold from 12 (model, prompt) pairs\\
across Maverick, Scout, Qwen $\times$ 4 prompt types
}} \\
$\downarrow$ \\
Vote vector $[v_1, v_2, v_3, v_4, v_5]$ \\
Margin $m = |2\sum v_i - 5|$ \\
$\downarrow$ \\
\framebox[0.9\columnwidth]{%
\parbox{0.85\columnwidth}{\centering
\textbf{Decision Module}\\
If $m \leq 1$: use Z3 verdict (if available)\\
Else: use ensemble majority
}} \\
$\downarrow$ \\
\textbf{Output:} Valid / Invalid \\
\hline
\end{tabular}
\caption{FregeLogic system architecture. The ensemble produces a vote vector; when the margin is low (3--2 split), the system defers to Z3 formal verification.}
\label{fig:architecture}
\end{figure}

Our system consists of three components: (1) an LLM ensemble that provides high-accuracy predictions through diverse model and prompt combinations, (2) a Z3-based formal verification pipeline that provides structurally-grounded logical judgments, and (3) a tiebreaker decision module that routes predictions to Z3 only when the ensemble produces a narrow vote margin (3--2 split). Figure~\ref{fig:architecture} provides an overview of the architecture.

\subsection{LLM Ensemble}

The ensemble consists of five classifiers, each defined by a (model, prompt) pair. We use three open-weight models accessed via Amazon Bedrock: Llama 4 Maverick (17B active parameters, MoE architecture), Llama 4 Scout (17B), and Qwen3 (32B). These models were selected to maximize architectural diversity (MoE vs.\ dense, two distinct model families) within a practical parameter budget (17B--32B active parameters), while remaining accessible through a single inference API. They are paired with four prompting strategies to create diversity across both model family and reasoning elicitation method:

\begin{itemize}
\item \textbf{Zero-shot:} A direct validity question with no examples.
\item \textbf{Few-shot:} Seven worked examples of valid and invalid syllogisms with explicit validity rules, balanced across plausibility conditions.
\item \textbf{Few-shot CoT:} Worked examples with step-by-step reasoning traces demonstrating structural analysis, including a believable-invalid example that explicitly names the content bias.
\item \textbf{Simple CoT:} A minimal prompt asking the model to identify logical structure before answering.
\end{itemize}

The five configurations are selected per fold based on combined score on a 200-sample inner subset (see Section~\ref{sec:setup}). The selection naturally balances model and prompt diversity, following the principle that ensemble accuracy benefits from uncorrelated errors across members \citep{bertolazzi2024systematic}.

For each syllogism, all five classifiers produce a binary prediction $v_i \in \{0, 1\}$, where 1 indicates valid. The ensemble majority vote is $\hat{y} = \mathbb{1}[\sum_i v_i > 2.5]$.

All models are called with temperature 0.0 to maximize output reproducibility. Responses are parsed using a multi-stage regex pipeline that checks for explicit answer patterns (e.g., ``ANSWER: true''), last-line heuristics, and fallback to the last occurrence of ``true'', ``false'', ``valid'', or ``invalid'' in the response. Parse failures default to invalid.

\subsection{Z3 Formal Verification}

The Z3 component \citep{demoura2008z3} provides structurally-grounded validity checking by encoding syllogisms in first-order logic (FOL) and testing satisfiability. We select Z3 over alternative provers for three reasons: (1) its native support for quantifiers and uninterpreted sorts allows direct encoding of syllogistic propositions without conversion to conjunctive normal form, unlike resolution-based provers such as Prover9 \citep{olausson2023linc}; (2) its mature Python API enables tight integration with our LLM pipeline; and (3) its proven effectiveness in neuro-symbolic NLP, where it has been used alongside other solvers for faithful logical reasoning \citep{pan2023logiclm, bayless2025neurosymbolic}. The pipeline consists of three stages.

\paragraph{Structure extraction.} An LLM is prompted to extract the logical structure of the syllogism as a JSON object containing the three terms, two premises, and conclusion, each annotated with its proposition type (A: universal affirmative, E: universal negative, I: particular affirmative, O: particular negative) and its subject and predicate terms. We use Amazon Bedrock's structured output API with a JSON schema that enforces well-formed output, reducing extraction failures from approximately 22\% (with free-form prompting) to near zero. Extraction is attempted with each of the three models in sequence (Maverick $\to$ Qwen $\to$ Scout, ordered by individual combined score), using the first successful parse. In practice, Maverick succeeds on 99.5--100\% of cases across folds.

\paragraph{FOL encoding.} Each proposition is encoded as a first-order formula over a declared sort \texttt{Thing} with unary predicates for each term. Based on diagnostic inspection of Felapton-type syllogisms in the dataset (all labelled valid), we adopt an Aristotelian interpretation with existential import, adding existence axioms ($\exists x: S(x)$) for subject terms in universal propositions:
\begin{itemize}
\item Type A (``All S are P''): $\forall x: S(x) \rightarrow P(x) \;\wedge\; \exists x: S(x)$
\item Type E (``No S are P''): $\forall x: S(x) \rightarrow \neg P(x) \;\wedge\; \exists x: S(x)$
\item Type I (``Some S are P''): $\exists x: S(x) \wedge P(x)$
\item Type O (``Some S are not P''): $\exists x: S(x) \wedge \neg P(x)$
\end{itemize}

\paragraph{Satisfiability check.} Validity is determined by a two-step procedure. First, we verify premise consistency: if $P_1 \wedge P_2$ is \textsc{unsat} (e.g., a Type~I premise paired with a contradicting Type~E premise over the same terms), the premises are mutually inconsistent and Z3 returns $\bot$, deferring to the ensemble. This guards against \textit{ex contradictione quodlibet} false positives, where any conclusion would be trivially entailed by inconsistent premises. Otherwise, validity is checked in the standard way:
\begin{equation}
\resizebox{0.9\columnwidth}{!}{$
\text{valid}(S) \iff (P_1 \wedge P_2\ \text{is SAT}) \wedge (P_1 \wedge P_2 \wedge \neg C\ \text{is UNSAT})
$}
\end{equation}

If the second check returns \textsc{unsat}, the syllogism is valid; if \textsc{sat}, it is invalid; if \textsc{unknown} (timeout at 5000ms), the result is undefined ($\bot$). When structure extraction fails or produces malformed output, Z3 also returns $\bot$.

The Z3 \textit{solver} is content-neutral by construction, since the formal encoding strips away all semantic content. However, the end-to-end pipeline's content independence is contingent on the extraction step. Our extraction failure disaggregation (Section~\ref{sec:results}) confirms that failure rates are uniform across plausibility subgroups. Pipeline accuracy depends entirely on correct structure extraction, which can fail when the LLM misidentifies proposition types or term boundaries.

\subsection{Tiebreaker Decision Fusion}

The decision module combines ensemble and Z3 predictions based on vote consensus, measured by the vote margin:
\begin{equation}
m(S) = |2 \cdot \textstyle\sum_i v_i - n|
\end{equation}
where $n = 5$ is the number of classifiers. For $n = 5$, the margin is 5 (unanimous), 3 (4--1 split), or 1 (3--2 split).

The final prediction is:
\begin{equation}
\resizebox{0.9\columnwidth}{!}{$
\text{pred}(S) = \begin{cases} z(S) & m(S) \leq 1 \wedge z(S) \neq \bot \\ \hat{y}(S) & \text{otherwise} \end{cases}
$}
\end{equation}
where $z(S)$ is the Z3 verdict and $\hat{y}(S)$ is the ensemble majority vote. The threshold $\tau = 1$ means Z3 is consulted only on 3--2 splits, which we hypothesize correspond to cases where content bias causes disagreement among the LLMs.

\section{Experimental Setup}
\label{sec:setup}

The dataset provided by the task organizers \citep{valentino-etal-2026-semeval} contains $N{=}960$ syllogisms annotated with validity labels and plausibility metadata, balanced across four subgroups: valid-believable (240), valid-unbelievable (240), invalid-believable (234), and invalid-unbelievable (246). No separate development set was provided.

We adopt a nested 5-fold cross-validation protocol to separate model selection from evaluation. In each outer fold, 768 syllogisms are used for calibration and 192 for evaluation. Within each calibration set, a 200-sample inner subset is randomly drawn for Phase~1 model selection, where all 12 combinations of 3 models and 4 prompts are scored. Two additional OpenAI models (GPT-OSS-120B and GPT-OSS-20B) were evaluated in preliminary experiments but excluded due to low combined scores (14.7--21.0; see Appendix~\ref{sec:excluded}). The top five configurations by combined score are selected per fold to form the ensemble. All LLM calls use Amazon Bedrock with temperature 0.0. The Z3 solver uses the Python API with a 5000ms timeout. Full prompts are in Appendix~\ref{sec:prompts}.

\paragraph{Existential import.} To determine the correct FOL encoding, we inspected the dataset for Darapti-type (AAI-3) and Felapton-type (EAO-3) syllogisms, whose validity depends on whether existential import is assumed. No Darapti candidates were found; seven Felapton candidates were found, all labelled valid. This confirms an Aristotelian interpretation, and we add existence axioms accordingly (Section~3.2).




\section{Results and Analysis}
\label{sec:results}

\subsection{Individual Model Performance}

Table~\ref{tab:individual} (Appendix~\ref{sec:individual_results}) presents
results for all 12 model-prompt combinations from a representative Phase~1
inner fold. Accuracy varies modestly across configurations (79.5--89.0\%),
but content effect varies substantially (3.49--13.71), driving large
differences in combined score. The best prompt strategy differs by model:
Maverick favours zero-shot and few-shot, while Qwen benefits from
chain-of-thought. Notably, Simple CoT achieves the lowest average content
effect (5.24) despite being the most under-specified prompt, suggesting that
minimal instructions may allow models to engage internal structural reasoning
rather than anchoring to example content. Chain-of-thought does not uniformly
reduce content effects; for Maverick, few-shot CoT increases CE from 3.83 to
10.11 compared to plain few-shot.

\subsection{Ensemble and Hybrid Strategies}

Table~\ref{tab:strategies} compares six fusion strategies aggregated across the five outer folds. The pure ensemble achieves 93.4\% accuracy but a content effect of 3.39. The Z3-only baseline achieves low accuracy (74.7\%) with a high content effect (26.28), driven by systematic over-prediction of invalidity.

\begin{table}[t]
\centering
\small
\begin{tabular}{lccc}
\hline
\textbf{Strategy} & \textbf{Acc} & \textbf{CE} & \textbf{Score} \\
\hline
Ensemble (pure) & 93.4$\pm$1.5 & 3.39$\pm$1.30 & 39.12$\pm$5.37 \\
+ Z3 Tiebreaker & \textbf{94.3$\pm$0.9} & \textbf{2.85$\pm$1.23} & \textbf{41.88$\pm$5.97} \\
+ Z3 Weighted & 93.4$\pm$1.5 & 3.39$\pm$1.30 & 39.12$\pm$5.37 \\
+ Z3 Veto & 74.1$\pm$2.1 & 26.70$\pm$2.97 & 17.18$\pm$0.84 \\
Confidence + Z3 & 91.7$\pm$1.1 & 6.15$\pm$2.00 & 31.77$\pm$3.82 \\
Top 3 + Z3 & 92.2$\pm$2.0 & 5.08$\pm$2.25 & 34.11$\pm$4.36 \\
\hline
Z3 Only & 74.7$\pm$2.0 & 26.28$\pm$2.88 & 17.39$\pm$0.87 \\
\hline
\end{tabular}
\caption{Comparison of fusion strategies (mean $\pm$ std across 5 outer folds). The tiebreaker strategy achieves the best combined score. Bold indicates the best result.}
\label{tab:strategies}
\end{table}

The tiebreaker strategy achieves the best combined score (41.88) by reducing the content effect from 3.39 to 2.85, a 16\% reduction, while also improving accuracy by 0.9 percentage points. This is consistent with our hypothesis that ensemble disagreements correspond to content-biased cases that Z3 can help correct.

Table~\ref{tab:subgroup} provides subgroup accuracy for the key strategies. The tiebreaker achieves more balanced accuracy across all four subgroups than either the pure ensemble or Z3 alone. The pure ensemble shows a gap between valid subgroups (95.9--96.0\%) and invalid subgroups (90.2--91.9\%); the tiebreaker narrows this gap (93.5--95.6\%), with the largest gains on invalid-believable cases (90.2\% $\to$ 94.5\%) where content bias is strongest.

\begin{table}[t]
\centering
\small
\begin{tabular}{lcccc}
\hline
\textbf{Strategy} & \textbf{VB} & \textbf{VU} & \textbf{IB} & \textbf{IU} \\
\hline
Ensemble (pure) & 95.9 & 96.0 & 90.2 & 91.9 \\
+ Z3 Tiebreaker & 95.6 & 93.8 & 94.5 & 93.5 \\
Z3 Only & 50.5 & 53.9 & 98.0 & 97.2 \\
\hline
\end{tabular}
\caption{Subgroup accuracy (\%) for key strategies, averaged across folds. VB = Valid-Believable, VU = Valid-Unbelievable, IB = Invalid-Believable, IU = Invalid-Unbelievable.}
\label{tab:subgroup}
\end{table}

The remaining strategies perform worse: Z3 veto (score 17.18) overrides correct high-consensus predictions, the weighted strategy never flips a majority, and the confidence-based strategy (91.7\%) trusts Z3 on both 3--2 and 4--1 splits (effectively lowering the margin threshold to $\tau{=}3$), overriding the ensemble on 37--52 cases per fold compared to 13--19 for the tiebreaker. Because Z3 accuracy on valid syllogisms is low (48.6\% on 3--2 split cases; see Section~\ref{sec:z3_invalidity}), extending Z3 authority to higher-consensus cases where the ensemble is typically correct degrades overall performance.

\subsection{Analysis of Tiebreaker Behavior}

Table~\ref{tab:tiebreaker} summarizes the tiebreaker mechanism's behavior aggregated across all five folds. The 3--2 vote split occurs in 7.9\% of cases (76 of 960). Z3 produces a usable verdict on all 76 cases, enabled by structured-output extraction which reduced failures to near zero. Of the 30 cases where Z3 overrides the ensemble majority, 19 are correct flips and 11 are wrong flips, yielding a net improvement of 8 correct predictions.

\begin{table}[t]
\centering
\small
\begin{tabular}{lrc}
\hline
\textbf{Metric} & \textbf{Count} & \textbf{\%} \\
\hline
Total evaluation instances & 960 & 100 \\
5--0 or 4--1 splits & 884 & 92.1 \\
3--2 splits (tiebreaker triggered) & 76 & 7.9 \\
\quad Z3 available on splits & 76 & 100.0 \\
\quad Degenerate premises & 1 & 0.1 \\
Z3 override decisions & 30 & \\
\quad Correct flips & 19 & \\
\quad Wrong flips & 11 & \\
\hline
\end{tabular}
\caption{Tiebreaker behavior aggregated across all five outer folds.}
\label{tab:tiebreaker}
\end{table}

When all five LLMs agree, the logical signal is typically strong enough to overcome content bias. The 3--2 splits indicate cases where some models are swayed by content while others are not; Z3 resolves these independently of semantic content. Coalition analysis reveals that Scout appears in the minority coalition on 53.9\% of 3--2 splits (41 of 76), above the 40\% expected by chance, suggesting greater susceptibility to content bias. Conversely, Maverick+FS appears in the minority least often (28.9\%, 22 of 76), well below the 40\% chance baseline, indicating the strongest resistance to content effects among the five classifiers.

\subsection{Z3 Invalidity Bias}
\label{sec:z3_invalidity}

The Z3-only baseline exhibits a pronounced invalidity bias: it achieves 97.6\% accuracy on invalid syllogisms (IB: 98.0\%, IU: 97.2\%) but only 52.2\% on valid ones (VB: 50.5\%, VU: 53.9\%). Because the Z3 solver itself is deterministically correct given well-formed input, this asymmetry originates in the extraction step. When the LLM misidentifies a proposition type or term boundary in a valid syllogism, the resulting FOL encoding almost always breaks the entailment chain, causing Z3 to correctly report unsatisfiability of the (now-corrupted) logical structure. Extraction errors on invalid syllogisms are less consequential: the syllogism is already non-entailing, so a different structural error is unlikely to accidentally produce a valid derivation. This directional asymmetry explains both the Z3-only subgroup collapse on valid cases and why the tiebreaker restricts Z3 authority to low-consensus cases, where the benefit of correcting content-biased ensemble errors outweighs the risk of extraction-induced false invalidity verdicts.

\subsection{Error Analysis}

Three main error sources remain: (1) \textbf{Z3 extraction errors}, where incorrect structure extraction (e.g., misidentified proposition types) produces wrong Z3 verdicts, accounting for 11 wrong flips across all folds; (2) \textbf{unanimous ensemble errors}, where all five LLMs agree incorrectly on syllogisms with strong content bias, preventing tiebreaker intervention; and (3) \textbf{degenerate premises}, where mutually inconsistent premises trigger the consistency check and Z3 defers to the ensemble (1 case across all folds).

Extraction failure disaggregation confirms that the structured-output pipeline is content-neutral: failure rates are 0.0\% for VB, IB, and IU subgroups, and 0.4\% for VU (1 of 241 cases), consistent with uniform distribution.

\begin{figure}[t]
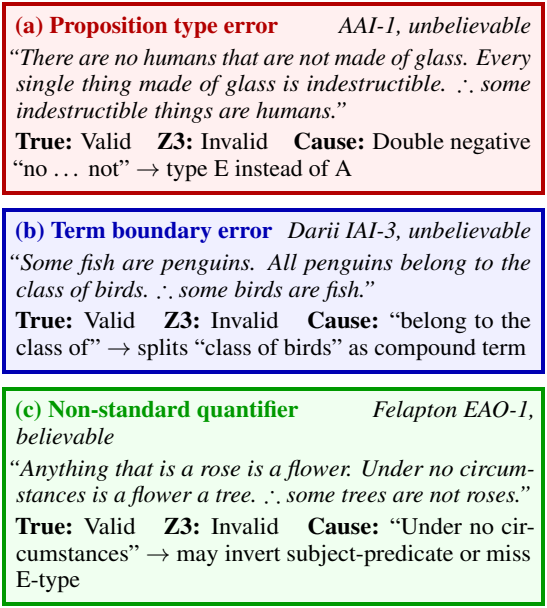

\centering
\small
\wrongflipbox{red!70!black}{red!6!white}{%
\textbf{\textcolor{red!70!black}{(a) Proposition type error}} \hfill \textit{AAI-1, unbelievable}\\[2pt]
\textit{``There are no humans that are not made of glass. Every single thing made of glass is indestructible. $\therefore$ some indestructible things are humans.''}\\[2pt]
\textbf{True:} Valid \;\; \textbf{Z3:} Invalid \;\; \textbf{Cause:} Double negative ``no \ldots\ not'' $\to$ type~E instead of~A
}\\[4pt]
\wrongflipbox{blue!70!black}{blue!6!white}{%
\textbf{\textcolor{blue!70!black}{(b) Term boundary error}} \hfill \textit{Darii IAI-3, unbelievable}\\[2pt]
\textit{``Some fish are penguins. All penguins belong to the class of birds. $\therefore$ some birds are fish.''}\\[2pt]
\textbf{True:} Valid \;\; \textbf{Z3:} Invalid \;\; \textbf{Cause:} ``belong to the class of'' $\to$ splits ``class of birds'' as compound term
}\\[4pt]
\wrongflipbox{green!60!black}{green!6!white}{%
\textbf{\textcolor{green!60!black}{(c) Non-standard quantifier}} \hfill \textit{Felapton EAO-1, believable}\\[2pt]
\textit{``Anything that is a rose is a flower. Under no circumstances is a flower a tree. $\therefore$ some trees are not roses.''}\\[2pt]
\textbf{True:} Valid \;\; \textbf{Z3:} Invalid \;\; \textbf{Cause:} ``Under no circumstances'' $\to$ may invert subject-predicate or miss E-type
}
\caption{Three representative wrong flips. All 11 go in the same direction: Z3 falsely rejects valid syllogisms due to extraction errors, not encoding errors.}
\label{fig:wrong_flips}
\end{figure}

All 11 wrong flips share the same direction: Z3 falsely rejects a valid syllogism. Eight involve unbelievable content, where absurd premises may compound extraction difficulty. Figure~\ref{fig:wrong_flips} presents three representative cases. In each, the Z3 solver produces a correct verdict given the (incorrectly) extracted structure, confirming that the bottleneck is extraction fidelity, not logical encoding.

Additional analysis of prompt strategies is in Appendix~\ref{sec:prompt_analysis}.

\section{Conclusion}

We presented FregeLogic, a hybrid neuro-symbolic system that combines an LLM ensemble with Z3 formal verification via a tiebreaker mechanism for syllogistic validity prediction. Our results support the hypothesis that ensemble disagreement signals content-biased predictions, which formal logic can help correct. In nested 5-fold cross-validation on the dataset ($N{=}960$), the tiebreaker achieves a combined score of 41.88, a 2.76-point improvement over the pure ensemble driven by a 16\% reduction in content effect and a 0.9\% accuracy gain. Adopting structured-output API calls for extraction reduced Z3 failure rates from approximately 22\% to near zero, and a two-step satisfiability check with Aristotelian existential import was validated against task annotations. Future work includes investigating whether the tiebreaker mechanism generalizes to other reasoning tasks and exploring adaptive thresholds for the vote margin cutoff.

\section*{Limitations}

While our approach involves no parameter updates, model and prompt selection, fusion strategy selection, and the tiebreaker threshold ($\tau{=}1$) were all tuned via nested cross-validation on the provided dataset; no independent test set was used. This selection procedure adds non-trivial setup complexity: each fold requires scoring all 12 (model, prompt) combinations on a 200-sample inner subset before the top-five ensemble is fixed, and the optimal configuration differs across folds (Appendix~\ref{sec:individual_results}), so the approach does not reduce to a single off-the-shelf recipe. The Z3 pipeline depends on LLM-based structure extraction, which, despite structured-output enforcement, can still produce semantically incorrect extractions (e.g., misidentified proposition types), accounting for the 11 wrong flips observed. Our system requires six LLM API calls per syllogism (five ensemble plus one extraction) and one Z3 solve, totalling approximately 4,300 tokens and 12.1 seconds of sequential latency per instance (2.8 seconds with parallel ensemble calls); Z3 solving adds under 10ms. Full cost profiling over 18,722 API calls is reported in Appendix~\ref{sec:cost}. We did not compare against larger single dense models (e.g., 70B+); whether architectural diversity at smaller scale outperforms a single larger model on this task remains an open question for future work.

\section*{Acknowledgments}

We thank the SemEval-2026 Task 11 organizers for designing a task that foregrounds the important challenge of content effects in reasoning evaluation.

\bibliography{custom}

\appendix

\section{Prompt Templates}
\label{sec:prompts}

We provide the four prompt templates used in our system. In each template, \texttt{\{syllogism\}} is replaced with the input syllogism text.

\subsection{Zero-Shot Prompt}

\begin{quote}
\small
\texttt{Determine if this syllogism is VALID.}

\texttt{VALID means: IF the premises were true, the conclusion MUST be true. Ignore whether premises are actually true in the real world.}

\texttt{Syllogism: \{syllogism\}}

\texttt{Answer with exactly one word: true or false}
\end{quote}

\subsection{Few-Shot Prompt}

\begin{quote}
\small
\texttt{Determine if this syllogism is VALID (conclusion necessarily follows from premises).}

\texttt{VALIDITY RULES:}\\
\texttt{- ``All A are B'' + ``All B are C'' -> ``All A are C'' (valid)}\\
\texttt{- ``No A are B'' + ``All C are A'' -> ``No C are B'' (valid)}\\
\texttt{- ``All A are B'' + ``Some C are A'' -> ``Some C are B'' (valid)}\\
\texttt{- ``All A are B'' + ``All C are B'' -> ``All A are C'' (invalid, undistributed middle)}\\
\texttt{- ``Some A are B'' does NOT guarantee ``All A are B''}

\texttt{EXAMPLES:}\\
\texttt{``All dogs are mammals. All mammals are animals. Therefore, all dogs are animals.'' -> true}\\
\texttt{``All birds are dinosaurs. All sparrows are birds. Therefore, all sparrows are dinosaurs.'' -> true}\\
\texttt{``No fish are mammals. All sharks are fish. Therefore, no sharks are mammals.'' -> true}\\
\texttt{``All reptiles are cold-blooded. Some lizards are reptiles. Therefore, some lizards are cold-blooded.'' -> true}\\
\texttt{``All lawyers are professionals. All doctors are professionals. Therefore, all lawyers are doctors.'' -> false}\\
\texttt{``Some politicians are corrupt. All senators are politicians. Therefore, some senators are corrupt.'' -> false}\\
\texttt{``All rocks are edible. Some clouds are rocks. Therefore, all clouds are edible.'' -> false}

\texttt{Syllogism: \{syllogism\}}

\texttt{Answer with exactly one word: true or false}
\end{quote}

\subsection{Few-Shot Chain-of-Thought Prompt}

\begin{quote}
\small
\texttt{Analyze this syllogism's logical VALIDITY.}

\texttt{IMPORTANT: VALID = conclusion MUST follow IF premises are assumed true. Ignore real-world facts.}

\texttt{RULES:}\\
\texttt{- ``All A are B'' + ``All B are C'' -> ``All A are C'' (valid chain)}\\
\texttt{- ``No A are B'' + ``All C are A'' -> ``No C are B'' (valid exclusion)}\\
\texttt{- ``All A are B'' + ``Some C are A'' -> ``Some C are B'' (Darii)}\\
\texttt{- ``All A are B'' + ``All C are B'' -> ``All A are C'' (invalid, undistributed middle)}\\
\texttt{- ``Some A are B'' means ONLY SOME, not all}

\texttt{WORKED EXAMPLES:}

\texttt{Example 1: ``All cats are mammals. All mammals are animals. Therefore, all cats are animals.''}\\
\texttt{- Structure: cats $\subseteq$ mammals $\subseteq$ animals}\\
\texttt{- Chain is complete. ANSWER: true}

\texttt{Example 2: ``All unicorns fly. All pegasi are unicorns. Therefore, all pegasi fly.''}\\
\texttt{- Premises are fantasy but structure is: pegasi $\subseteq$ unicorns $\subseteq$ fly}\\
\texttt{- Valid chain regardless of real-world truth. ANSWER: true}

\texttt{Example 3: ``All athletes are healthy. All healthy people exercise. Therefore, all athletes exercise.''}\\
\texttt{- Chain: athletes -> healthy -> exercise}\\
\texttt{- Chain is complete. ANSWER: true}

\texttt{Example 4: ``All doctors are professionals. All lawyers are professionals. Therefore, all doctors are lawyers.''}\\
\texttt{- ``Professionals'' appears as PREDICATE in both premises.}\\
\texttt{- Middle term is undistributed: we only know both are subsets of professionals, not that they overlap.}\\
\texttt{- Despite the believable surface, the structure is invalid. ANSWER: false}

\texttt{Example 5: ``All cats are pets. All dogs are pets. Therefore, all cats are dogs.''}\\
\texttt{- Both subsets of pets, but could be separate}\\
\texttt{- Undistributed middle. ANSWER: false}

\texttt{Example 6: ``Some birds can fly. All penguins are birds. Therefore, some penguins can fly.''}\\
\texttt{- ``Some birds'' doesn't tell us WHICH birds}\\
\texttt{- Cannot guarantee any penguin is in the flying subset. ANSWER: false}

\texttt{Syllogism: \{syllogism\}}

\texttt{Think through the structure briefly, then write your final answer as: ANSWER: true or ANSWER: false}
\end{quote}

\subsection{Simple Chain-of-Thought Prompt}

\begin{quote}
\small
\texttt{Is this syllogism logically VALID? (If premises were true, must conclusion be true?)}

\texttt{Syllogism: \{syllogism\}}

\texttt{First, identify the logical structure. Then determine if the conclusion necessarily follows.}

\texttt{End your response with exactly: ANSWER: true or ANSWER: false}
\end{quote}

\section{Z3 Structure Extraction Prompt}
\label{sec:z3prompt}

The following prompt is used to extract syllogistic structure for Z3 encoding. When supported, extraction uses Bedrock's structured output API with a JSON schema enforcing the expected format; otherwise, the model's free-form response is parsed.

\begin{quote}
\small
\texttt{Extract the logical structure of this syllogism.}

\texttt{SYLLOGISM: \{syllogism\}}

\texttt{Proposition types:}\\
\texttt{- A: ``All S are P'' / ``Every S is P''}\\
\texttt{- E: ``No S are P''}\\
\texttt{- I: ``Some S are P'' / ``At least one S is P''}\\
\texttt{- O: ``Some S are not P''}

\texttt{The CONCLUSION follows ``therefore/hence/thus/consequently/so''.}

\texttt{Output ONLY this JSON (replace t1/t2/t3 with the exact term WORDS from the syllogism text):}\\
\texttt{\{"terms": ["t1", "t2", "t3"],}\\
\texttt{"premise1": \{"type": "A/E/I/O",}\\
\texttt{\quad"subject": "term <- exact word(s) from text",}\\
\texttt{\quad"predicate": "term <- exact word(s) from text"\},}\\
\texttt{"premise2": \{"type": "A/E/I/O",}\\
\texttt{\quad"subject": "term <- exact word(s) from text",}\\
\texttt{\quad"predicate": "term <- exact word(s) from text"\},}\\
\texttt{"conclusion": \{"type": "A/E/I/O",}\\
\texttt{\quad"subject": "term <- exact word(s) from text",}\\
\texttt{\quad"predicate": "term <- exact word(s) from text"\}\}}
\end{quote}

\section{Excluded Models}
\label{sec:excluded}

Table~\ref{tab:excluded} reports results for GPT-OSS-120B and GPT-OSS-20B, which were evaluated but excluded from the final ensemble due to low combined scores driven by high content effects.

\begin{table}[t]
\centering
\small
\begin{tabular}{llccc}
\hline
\textbf{Model} & \textbf{Prompt} & \textbf{Acc} & \textbf{CE} & \textbf{Score} \\
\hline
GPT-120B & Zero-Shot & 80.0 & 15.71 & 20.96 \\
GPT-120B & Few-Shot & 78.0 & 18.64 & 19.61 \\
GPT-120B & FS-CoT & 76.0 & 20.32 & 18.72 \\
GPT-120B & Simple CoT & 75.5 & 19.88 & 18.69 \\
\hline
GPT-20B & Zero-Shot & 79.0 & 13.67 & 21.43 \\
GPT-20B & Few-Shot & 78.5 & 14.71 & 20.91 \\
GPT-20B & FS-CoT & 73.0 & 18.92 & 18.29 \\
GPT-20B & Simple CoT & 65.5 & 31.14 & 14.65 \\
\hline
\end{tabular}
\caption{Results for excluded models on a 200-sample evaluation set. These models exhibited higher content effects on average than the selected models, particularly for chain-of-thought prompts, resulting in lower combined scores across all four prompt types.}
\label{tab:excluded}
\end{table}

\section{Individual Model Results}
\label{sec:individual_results}

Table~\ref{tab:individual} presents results for all 12 model-prompt configurations from a representative Phase~1 inner fold (200 samples).

\begin{table}[t]
\centering
\small
\begin{tabular}{llccc}
\hline
\textbf{Model} & \textbf{Prompt} & \textbf{Acc} & \textbf{CE} & \textbf{Score} \\
\hline
Maverick & Zero-Shot & 86.0 & 3.49 & 34.38 \\
Maverick & Few-Shot & 88.0 & 3.83 & 34.17 \\
Maverick & FS-CoT & 88.5 & 10.11 & 25.97 \\
Maverick & Simple CoT & 84.0 & 5.86 & 28.71 \\
\hline
Scout & Zero-Shot & 81.0 & 13.71 & 21.96 \\
Scout & Few-Shot & 89.0 & 6.92 & 29.00 \\
Scout & FS-CoT & 88.5 & 11.21 & 25.27 \\
Scout & Simple CoT & 85.0 & 4.79 & 30.84 \\
\hline
Qwen & Zero-Shot & 80.5 & 8.86 & 24.48 \\
Qwen & Few-Shot & 79.5 & 11.37 & 22.62 \\
Qwen & FS-CoT & 89.0 & 5.20 & 31.52 \\
Qwen & Simple CoT & 88.0 & 5.07 & 31.39 \\
\hline
\end{tabular}
\caption{Individual model-prompt results from a representative Phase~1 inner fold (200 samples). Acc = Accuracy (\%), CE = Content Effect, Score = Combined Score. FS-CoT = Few-Shot Chain-of-Thought.}
\label{tab:individual}
\end{table}

\section{Prompt Strategy Analysis}
\label{sec:prompt_analysis}

Table~\ref{tab:prompt_avg} shows average metrics by prompt type across the three selected models on a representative 200-sample inner fold. Simple CoT achieves the best average combined score (30.31) due to its low content effect (5.24), despite not having the highest accuracy. Few-shot CoT increases content effect for Maverick and Scout while decreasing it for Qwen, indicating an interaction between model architecture and reasoning elicitation. This suggests that minimal reasoning prompts may be preferable to elaborate demonstrations when the goal is content-independent reasoning.

\begin{table}[!h]
\centering
\small
\begin{tabular}{lccc}
\hline
\textbf{Prompt Type} & \textbf{Avg Acc} & \textbf{Avg CE} & \textbf{Avg Score} \\
\hline
Zero-Shot & 82.50 & 8.69 & 26.94 \\
Few-Shot & 85.50 & 7.37 & 28.60 \\
Few-Shot CoT & 88.67 & 8.84 & 27.59 \\
Simple CoT & 85.67 & 5.24 & 30.31 \\
\hline
\end{tabular}
\caption{Average metrics by prompt type across three models on a representative 200-sample inner fold.}
\label{tab:prompt_avg}
\end{table}

\section{Strategy Trade-offs}
\label{sec:plots}

\begin{figure}[!h]
\centering
\includegraphics[width=1\columnwidth]{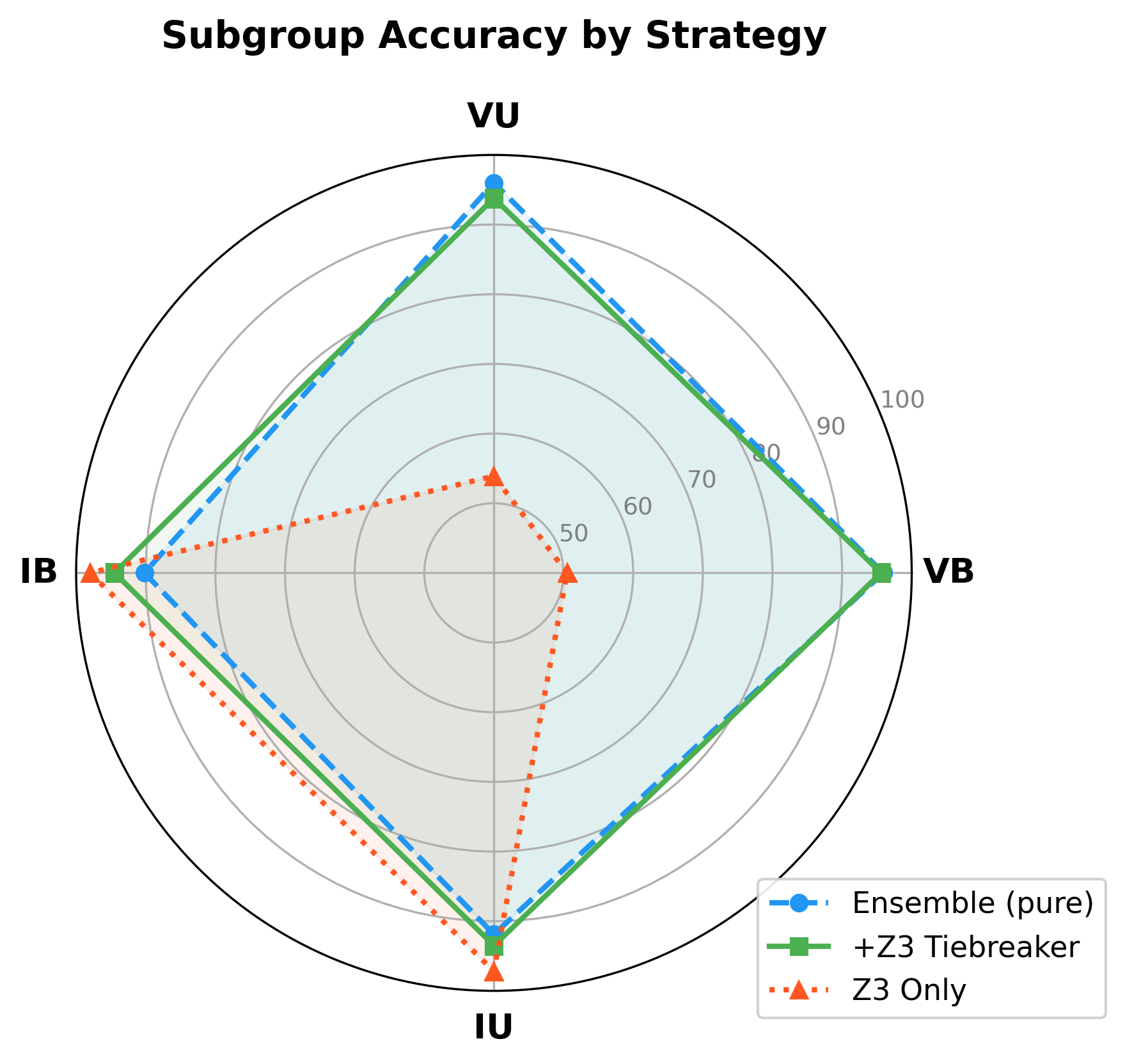}
\caption{Subgroup accuracy by strategy. The tiebreaker achieves the most balanced profile across all four subgroups.}
\label{fig:radar}
\end{figure}

Figure~\ref{fig:radar} shows subgroup accuracy for the three key strategies. The tiebreaker (solid green) achieves the most symmetric profile, while the pure ensemble (dashed blue) favours valid subgroups and Z3 Only (dotted red) favours invalid subgroups.

\section{Computational Cost}
\label{sec:cost}

\begin{figure*}[t]
\centering
\includegraphics[width=\textwidth]{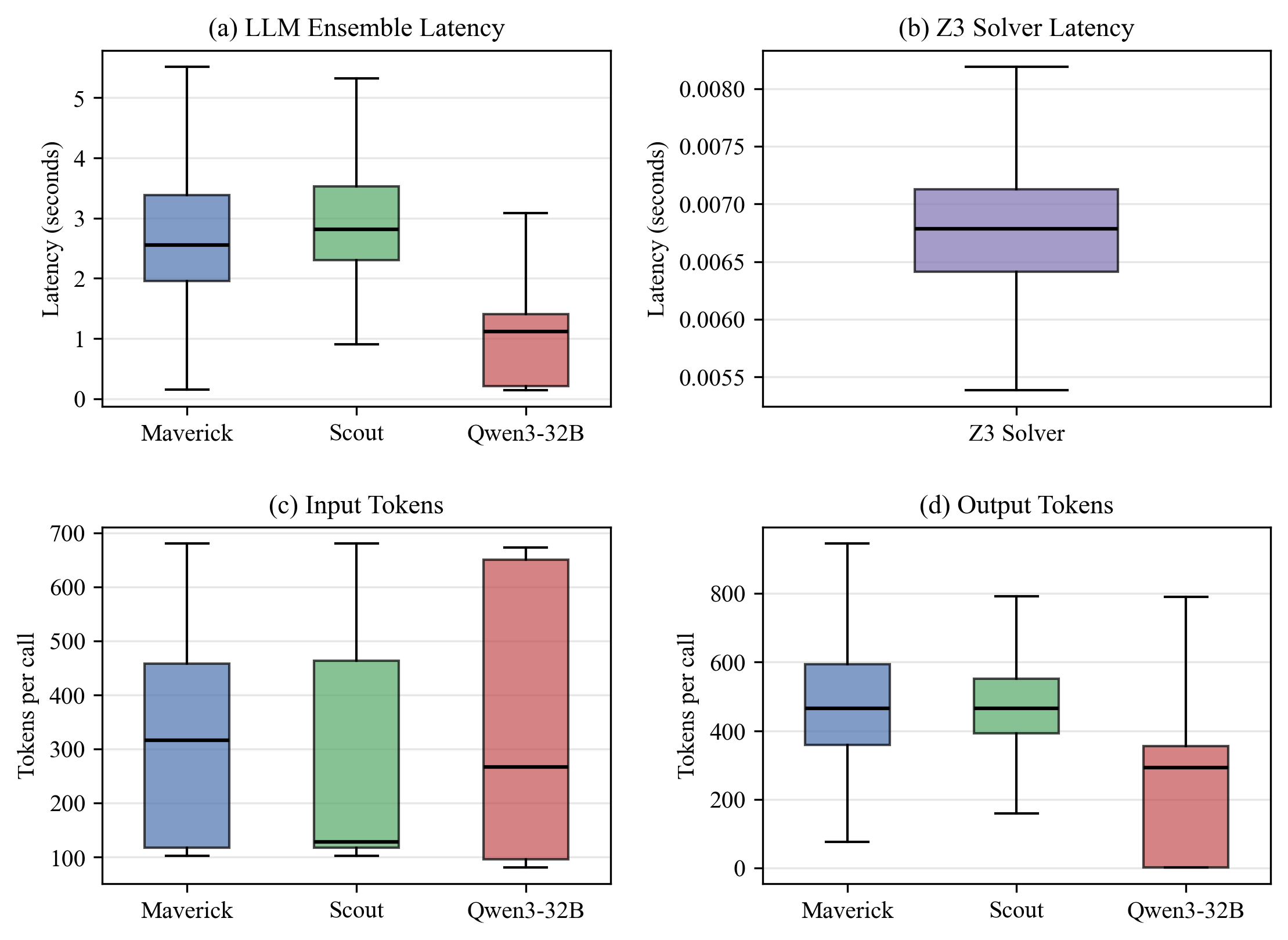}
\caption{Per-call cost distributions across the full cross-validation run (18,722 calls). (a)~LLM ensemble latency, (b)~Z3 solver latency (note the millisecond scale), (c)~input tokens per call, (d)~output tokens per call. Outliers suppressed for readability. The computational bottleneck is LLM inference, not formal verification.}
\label{fig:cost_latency}
\end{figure*}

Table~\ref{tab:cost} reports per-component cost profiling from the full nested cross-validation run (18,722 API calls across model selection and evaluation phases). Qwen3-32B is the fastest model per call (median 1.1s), while Maverick and Scout average 2.5--2.8s at the median. Z3 solving is negligible: 960 calls complete in 8.5 seconds total (median 7ms). At inference time, each syllogism requires five ensemble calls, one extraction call, and one Z3 solve, consuming approximately 2,200 input tokens and 2,100 output tokens.

\begin{table}[t]
\centering
\small
\begin{tabular}{lrccc}
\hline
\textbf{Component} & \textbf{$n$} & \textbf{p50 (s)} & \textbf{p95 (s)} & \textbf{Tokens} \\
\hline
Maverick (ens.) & 7,264 & 2.55 & 5.25 & 5.84M \\
Scout (ens.) & 4,576 & 2.82 & 5.26 & 3.69M \\
Qwen3-32B (ens.) & 5,920 & 1.11 & 1.86 & 3.35M \\
Qwen3-32B (extr.) & 2 & 0.49 & 0.49 & 1.4K \\
Z3 Solver & 960 & 0.007 & 0.009 & --- \\
\hline
\textbf{Total} & \textbf{18,722} & & & \textbf{12.87M} \\
\hline
\end{tabular}
\caption{Per-component cost profiling across the full nested 5-fold cross-validation run. $n$ = number of calls, p50/p95 = median and 95th-percentile latency, Tokens = total input + output tokens. Extraction calls are rare because Maverick succeeds on $>$99.5\% of cases.}
\label{tab:cost}
\end{table}

Figure~\ref{fig:cost_latency} shows per-call distributions across all four dimensions. LLM ensemble calls range from 1 to 5 seconds, with Qwen3-32B consistently faster than the Llama~4 models but consuming more input tokens due to longer prompt encodings. Maverick and Scout produce more output tokens on average (488 and 480 vs.\ 228 for Qwen), reflecting more verbose chain-of-thought responses. Z3 solve times are three orders of magnitude smaller than any LLM call, confirming that the formal verification step adds negligible overhead.

\end{document}